\documentclass[10pt,twocolumn,letterpaper]{article}

\usepackage{iccv}
\usepackage{times}
\usepackage{epsfig}
\usepackage{graphicx}
\usepackage{amsmath}
\usepackage{amssymb}
\usepackage[linesnumbered,ruled]{algorithm2e}
\usepackage{algpseudocode}
\usepackage{bm}
\usepackage{multirow}
\usepackage{diagbox}
\usepackage{array}
\usepackage{booktabs}
\usepackage{nth}
\usepackage{nicefrac}
\usepackage{relsize}
\usepackage{xspace} 
\usepackage{bm}
\usepackage{enumerate}
\usepackage{subcaption}
\usepackage{enumitem}
\usepackage{array}
\newcommand{\arl}{ARL}
\newcommand{\methiccv}{TIMAM}
\DeclareMathOperator*{\E}{\mathbb{E}}
\newcolumntype{P}[1]{>{\centering\arraybackslash}p{#1}}
\newcommand{\mytilde}{\raise.17ex\hbox{$\scriptstyle\mathtt{\sim}$}}

\usepackage[pagebackref=true,breaklinks=true,colorlinks,bookmarks=false]{hyperref}

\iccvfinalcopy

\def\iccvPaperID{60}

\ificcvfinal\fi
\begin{document}
	
	\title{Adversarial Representation Learning for Text-to-Image Matching}
	\author{Nikolaos Sarafianos \quad \quad \quad Xiang Xu \quad \quad \quad Ioannis A. Kakadiaris\\
		Computational Biomedicine Lab\\
		University of Houston\\
		{\tt\small \{nsarafianos, xxu21\}@uh.edu, ikakadia@central.uh.edu}
	}
	
	\maketitle

	\begin{abstract}
		For many computer vision applications such as image captioning, visual question answering, and person search, learning discriminative feature representations at both image and text level is an essential yet challenging problem. Its challenges originate from the large word variance in the text domain as well as the difficulty of accurately measuring the distance between the features of the two modalities. Most prior work focuses on the latter challenge, by introducing loss functions that help the network learn better feature representations but fail to account for the complexity of the textual input. With that in mind, we introduce TIMAM: a \textit{Text-Image Modality Adversarial Matching} approach that learns modality-invariant feature representations using adversarial and cross-modal matching objectives. In addition, we demonstrate that BERT, a publicly-available language model that extracts word embeddings, can successfully be applied in the text-to-image matching domain. The proposed approach achieves state-of-the-art cross-modal matching performance on four widely-used publicly-available datasets resulting in absolute improvements ranging from 2\% to 5\% in terms of rank-1 accuracy. 
	\end{abstract}
	\vspace{-0.5cm}
	\section{Introduction}
	We set out to develop a cross-modal matching method that given a textual description, identifies and retrieves the most relevant images. For example, given the sentence ``\textit{A woman with a white shirt carrying a black purse on her hand}'' we aspire to obtain images of individuals with such visual attributes. The first challenge of matching images and text is the large word variability in the textual descriptions even when they are describing the same image. What is considered as important information for one is not necessarily the same for another annotator. At the same time, textual descriptions might contain mistakes, the descriptions can be too long or the annotator might describe additional information that is available on the image but is not related to the primary point of interest (\eg, human, object). All of these factors make text-to-image matching a difficult problem since learning good feature representations from such descriptions is not straightforward.

	\begin{figure}[t]
		\centering
		\includegraphics[width=0.97\linewidth]{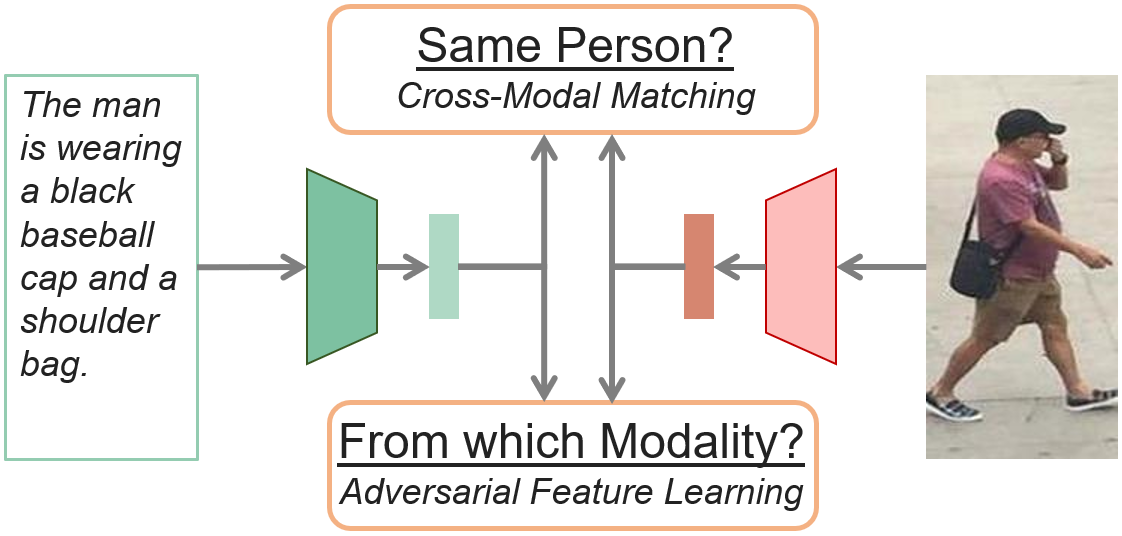}
		\caption{We learn discriminative embeddings from the visual and textual inputs by: (i) matching the distributions of the features that belong to the same identity, and (ii) employing a modality discriminator that tries to distinguish the encoded textual from the visual examples.}
		\label{fig:framework}
		\vspace{-0.45cm}
	\end{figure}
	
	A second major challenge of text-to-image matching is how to accurately measure the distance between the text and image features. During deployment, the distance between the probe text features and all the gallery image features is computed and the results are ranked based on this criterion. Most existing methods introduce loss functions to tackle this challenge. For example, Li \etal~\cite{li2017identity} proposed to ``bring close-together'' cross-modal features originating from the same identity and ``push away'' features from different identities. Although such methods have continuously outperformed previous state-of-the-art, their performance remains unsatisfactory. For example, the best performing text-to-image matching method on the CUHK-PEDES dataset~\cite{li2017person} is below 50\% in terms of rank-1 accuracy. Finally, most methods usually rely on some assumptions under which they perform matching. For example, in the work of Chen \etal~\cite{chen2018improving}, part-of-speech (POS) tagging is performed to extract local phrases (\eg, nouns with adjectives). However, when we performed POS tagging on the same textual inputs, we observed that important information is lost since the same word can be tagged differently depending on the context or its location within the sentence (\eg, the word ``\textit{t-shirt}'' was frequently identified as an adjective although it was used as a noun in the description). 
	
	In this paper, our objectives are: to (i) learn discriminative representations from both the visual and the textual inputs; and (ii) improve upon previous text-to-image matching methods in terms of how the word embeddings are learned. To accomplish these tasks, we introduce TIMAM: a \textit{Text-Image Modality Adversarial Matching} method that performs matching between the two modalities and achieves state-of-the-art results without requiring any additional supervision. 
	
	The first contribution of this work is an adversarial representation learning (\arl) framework that brings the features from both modalities ``close-to-each-other''. The textual and visual feature representations are fed to a discriminator that aims to identify whether the input is originating from the visual or textual modality. 
	By learning to fool the discriminator, we can learn modality-invariant feature representations that are capable of successfully performing text-to-image matching. The adversarial loss of the discriminator along with an identification loss and a cross-modal projection matching loss are used to jointly train the whole network end-to-end. We demonstrate that adversarial learning is well-suited for cross-modal matching and that it results in improved rank-1 accuracy. Our second contribution stems from improving upon previous text-to-image matching methods in terms of how the word embeddings are learned. We borrow from the NLP community a recent language representation model named BERT~\cite{devlin2018bert}, which stands for \textit{Bidirectional Encoder Representations from Transformers}. We demonstrate that such a model can successfully be applied to text-to-image matching and can significantly improve the performance of existing approaches. Each description is fed to the language model which extracts word representations that are then fed to an LSTM and mapped to a final sentence embedding. 
	
	Thus, TIMAM results in more discriminative feature representations learned from the proposed objective functions, while using the learning capabilities of the backbones of the two modalities. Through experiments, ablation studies and qualitative results, we demonstrate that:
	\begin{itemize}[noitemsep]
		\item[--] Adversarial learning is well-suited for cross-modal matching and that it results in more discriminative embeddings from both modalities. Using our proposed learning approach, we observe improvements ranging from 2\% to 5\% in terms of rank-1 accuracy over the previous best-performing techniques.
		\item[--] Pre-trained language models can successfully be applied to cross-modal matching. By leveraging the fine-tuning capabilities of BERT, we learn better word embeddings. Our experimental results indicate rank-1 accuracy improvements ranging from 3\% to 5\% over previous work when features are learned in this manner.
	\end{itemize}
	
	\section{Related Work}
	\noindent\textbf{Text-Image Matching}: Learning cross-modal embeddings has numerous applications~\cite{xu2019multilevel,zhang2018man} ranging from PINs using facial and voice information~\cite{nagrani2018learnable}, to generative feature learning~\cite{gu2018look} and domain adaptation\cite{xu2019deep,xu2019d}. Nagrani \etal~\cite{nagrani2018learnable} demonstrated that a joint representation can be learned from facial and voice information and introduced a curriculum learning strategy~\cite{bengio2009curriculum,sarafianos2017curriculum,sarafianos2018curriculum} to perform hard negative mining during training. Text-to-image matching is a well-studied problem in computer vision~\cite{cao2016deep, huang2019bi, li2018self, nam2017dual, sivic2003video, song2019polysemous, wang2016learning, wang2018joint, yan2015deep, zhang2018cross} facilitated by datasets describing objects, birds, or flowers~\cite{lin2014microsoft, reed2016learning, young2014image}. A relatively new application of text-to-image matching is person search the task of which is to retrieve the most relevant frames of an individual given a textual description as an input. Most methods~\cite{jing2018cascade, li2017identity, li2017person} rely on a relatively similar procedure: (i) extract discriminative image features using a deep neural network, (ii) extract text features using an LSTM, and (iii) propose a loss function that measures as accurately as possible the distance between the two embeddings. To improve the performance, some interesting ideas include jointly learning the 2D pose along with attention masks~\cite{jing2018cascade} or associating image features from sub-regions with the corresponding phrases in the text~\cite{chen2018improving}. 
	While such approaches have shown significant improvements, they: (i) ignore the large variability of the textual input by solely relying on an LSTM to model the input sentences, and (ii) demonstrate unsatisfactory results as the best text-to-image rank-1 accuracy results in the literature are below 50\% and 40\% in the CUHK-PEDES~\cite{li2017person} and Flickr30K~\cite{plummer2015flickr30k} datasets, respectively. Finally, there have been a few works recently~\cite{gu2018look,he2017unsupervised,peng2019cm,wang2017adversarial,zhu2019r2gan} that have applied adversarial learning in cross-modal matching applications. Zhu \etal~\cite{zhu2019r2gan} introduced \(R^2GAN\): a text-to-image matching method accompanied by a large-scale dataset to perform retrieval of food recipes. Unlike the rest of the datasets where the textual input is a description of the image, in this case the input consists of a title, a set of ingredients and a list of instructions for the recipe, which introduce additional challenges on how to properly handle and learn discriminative textual representations. 
	
	\begin{figure*}[t]
		\centering
		\includegraphics[width=0.99\linewidth]{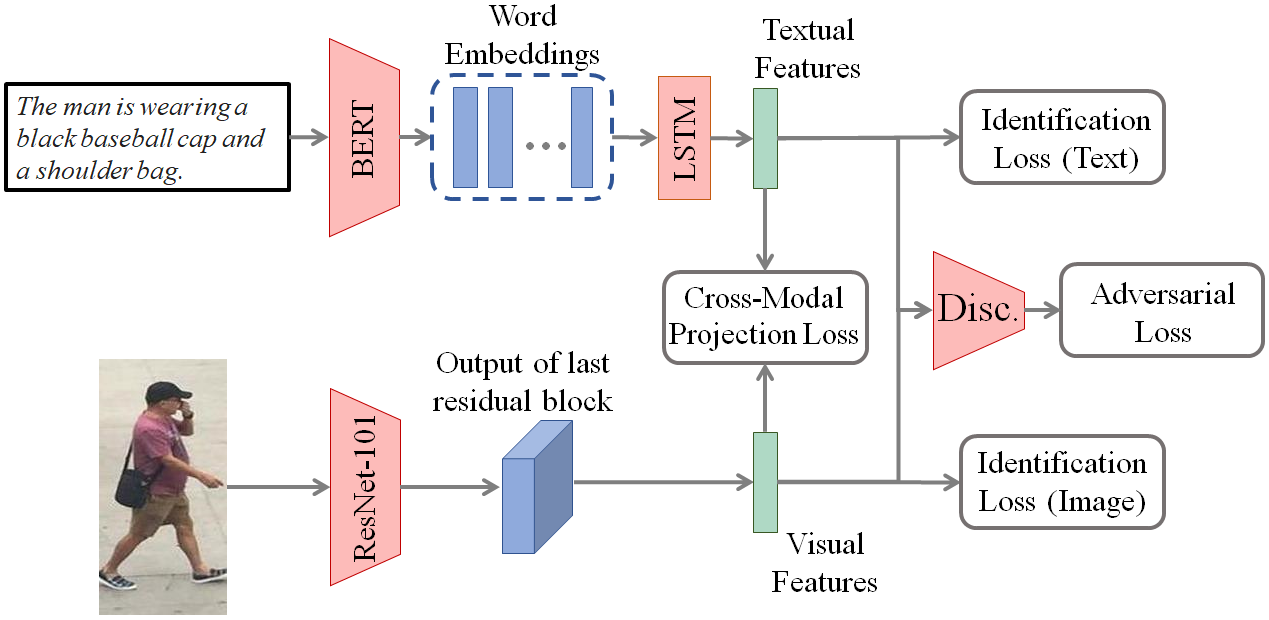}
		\caption{TIMAM consists of three modules: (i) the feature extraction module which extracts textual and visual features using their corresponding backbone architectures, (ii)  the identification and cross-modal projection losses that match the feature distributions originating from the same identity, and (iii) an adversarial discriminator that pushes the model to learn modality-invariant representations for effective text-image matching.}
		\vspace{-0.4cm}
		\label{fig:arch}
	\end{figure*}

	\noindent\textbf{Overview of BERT}: BERT~\cite{devlin2018bert} is a deep language model capable of extracting discriminative embeddings. Unlike previous approaches~\cite{peters2018deep, radford2018} that processed the input sequences either from left to right or combined left-to-right and right-to-left training, BERT relies on applying the bidirectional training of Transformer~\cite{vaswani2017attention} to language modeling. By leveraging the Transformer architecture (which is an attention mechanism) BERT learns the contextual relations between the words in a textual description. In addition, it introduces a word masking mechanism, which masks 15\% of the words with a token and then a model is trained to predict the original value of the masked words based on the context. In that way, BERT learns robust word embeddings that can be fine-tuned for a wide range of tasks.

	\section{Methodology}
	In this section, we introduce TIMAM: a cross-modal matching approach that learns to match the feature representations from the two modalities in order to perform both text-to-image and image-to-text retrieval.
	\subsection{Joint Feature Learning}\label{ssec:feat}
	During training, our objective is to learn discriminative visual and textual feature representations capable of accurately retrieving the ID (or the category) of the input from another modality. The training procedure is depicted in Figure~\ref{fig:arch} and is described in detail below. Specifically, our input at training-time consists of triplets \((V_i,T_i,Y_i)\) where \(V_i\) is the image input from the visual domain \(V\), \(T_i\) a textual description from the textual domain \(T\) describing that image, and \(Y_i\) is the identity/category of the input. To learn the visual representations denoted by \(\phi(V_i)\), any image classification model can be used as a backbone network (a ResNet-101 network is used in this work). The feature map of the last residual block is projected to the dimensionality of the feature vector using global average pooling and a fully-connected layer. We opted for the original backbone architecture without any attention blocks~\cite{chen2018improving, sarafianos2018deep} in order to keep the backbones simple and easy-to-reproduce in any framework and to avoid having to learn more parameters. 
	
	Learning discriminative representations from both modalities is of paramount importance for text-to-image matching. While for the image domain, most existing methods~\cite{chen2018improving, jing2018cascade, li2017identity, zheng2017dual} rely on deep architectures that have demonstrated their capability of extracting discriminative features for a wide range of tasks, this is not the case for the text domain. Prior work usually relies on a single LSTM~\cite{hochreiter1997long} to model the textual input and learn the features that correspond to the input sentence. We argue that one of the main reasons that prevent existing computer vision methods from performing well on text-to-image matching problems is due to the fact that the textual features are not discriminative enough. To address this limitation, we borrow from the NLP community a recently proposed language representation model named BERT. The sequence of word embeddings extracted from BERT is then fed to a bidirectional LSTM~\cite{hochreiter1997long}, which effectively summarizes the content of the input textual description. Finally, the textual representation denoted by \(\tau(T_i)\) is obtained by projecting the output of the LSTM to the dimensionality of the feature vector using a fully-connected layer. The reason an LSTM is employed on the output word embeddings is because it gives us the flexibility to initially ``freeze'' the weights of the language model and fine-tune only the LSTM along with the fully-connected layer and thus, significantly reducing the number of parameters. Once an adequate performance is observed, we ``unfreeze'' the weights of the language model and the whole network is trained end-to-end. 
	
	\subsection{Cross-Modal Matching}
	Given the visual and textual features, our aim is to introduce loss functions that will bring the features originating from the same identity/category close together and push away features originating from different identities. To accomplish this task, we introduce two loss functions for identification and cross-modal matching. The identification loss is a norm-softmax cross entropy loss which is commonly used in face identification applications~\cite{liu2017sphereface, wang2017normface, xu2017evaluation} that introduces an \(L_2\)-normalization on the weights of the output layer. By doing so, it enforces the model to focus on the angle between the weights of the different samples instead of their magnitude. For the visual features, the norm-softmax cross entropy loss can be described as follows: 
	\begin{equation}\label{eq:Liv}
	\begin{aligned}
	L_{I}^{V} &= -\frac{1}{B}\sum_{i=1}^{B}\log\left(\frac{\exp(W_i^T\phi(V_i) + b_i )}{\sum_{j}\exp(W_j^T\phi(V_i) + b_j )}\right), \\ &\quad\quad\quad\quad s.t. \; ||W_j||=1, \; \forall j \in [1,B]\;,
	\end{aligned}
	\end{equation}
	where \(I\) stands for identification, \(V\) corresponds to the visual modality, \(B\) is the batch size, and \(W_i, b_i\) are the weights and the bias of the classification layer for the visual feature representation \(\phi(V_i)\).
	The loss for the textual features \(L_{I}^T\) is computed in a similar manner and the final classification loss for identification \(L_{I} = L_{I}^V + L_{I}^T \). It is worth noting that for datasets that do not have ID labels but only image-text pairs (\eg, the Flickr30K dataset~\cite{plummer2015flickr30k}), we assign a unique ID to each image and use that ID as ground-truth for the identification loss. However, focusing solely on performing accurate identification is not sufficient for cross-modal matching since no association between the representations of the two modalities has been introduced thus far. To address this challenge, we use the cross-modal projection matching loss~\cite{zhang2018deep} which incorporates the cross-modal projection into the KL divergence measure to associate the representations across different modalities. The text representation is first normalized \(\bar{\tau}(T_j) = \frac{\tau(T_j)}{||\tau(T_j)||}\) and then the probability of matching \(\phi(V_i)\) to \(\bar{\tau}(T_j)\) is given by: 
	\begin{equation}\label{eq:pij}
	p_{i,j} = \frac{\exp\left(\phi(V_i)^T \bar{\tau}(T_j)\right)}{\sum_{k=1}^{B}\exp\big(\phi(V_i)^T\bar{\tau}(T_k)\big)} \;. 
	\end{equation}
	The multiplication between the transposed image embedding and the normalized textual embedding reflects the scalar projection between \(\phi(V_i)\) onto \(\bar{\tau}(T_j)\), while the probability \(p_{i,j}\) represents the proportion of this scalar projection among all scalar projections between pairs in a batch. Thus, the more similar the image embedding is to the textual embedding, the larger the scalar projection is from the former to the latter. 
	Since in each mini-batch there might be more than one positive matches (\ie, visual and textual features originating from the same identity) the true matching probability is normalized as follows: 
	\(q_{i,j} = \nicefrac{Y_{i,j}}{\sum_{k=1}^{B}(Y_{i,k})}\).
	The cross-modal projection matching loss of associating \(\phi(V_i)\) with the correctly matched text features is then defined as the KL divergence from the true matching distribution \(q_i\) to the probability of matching \(p_i\). For each batch this loss is defined as: 
	\begin{equation}\label{eq:cmpm}
	L_M^V = -\frac{1}{B}\sum_{i=1}^{B}\sum_{j=1}^{B}p_{i,j}\log\left(\frac{p_{i,j}}{q_{i,j}+\epsilon} \right), 
	\end{equation}
	where \(M\) denotes matching, and \(\epsilon\) is a very small number for preventing division by zero. The same procedure is followed to perform the opposite matching (\ie, from text to image) to compute loss \(L_M^T\) during which the visual features are normalized instead of \(\tau(T_i)\) to compute Eq.~(\ref{eq:pij}). Finally, the summation of the two individual losses constitutes the cross-modal projection matching loss \(L_M = L_M^V + L_M^T\). 
	
	\subsection{Adversarial Cross-Modal Learning}
	When training adversarial neural networks~\cite{cao2018partial, ganin2014unsupervised, tzeng2015simultaneous} a two-player minimax game is played between a discriminator \(D\) and a feature generator \(G\). Both \(G\) and \(D\) are jointly trained so as \(G\) tries to fool \(D\) and \(D\) tries to make accurate predictions. For the text-to-image matching problem, the two backbone architectures discussed in Section~\ref{ssec:feat} serve as the feature generators \(G^V\) and \(G^T\) for the visual and textual modalities that produce feature representations \(\phi(V_i)\) and \(\tau(T_i)\), respectively. The key idea is to learn a good general representation for each input modality that maximizes the matching performance, yet obscures the modality information. By learning to fool the modality discriminator, better feature representations are learned, that are capable of performing text-to-image matching. The generated embeddings are fed to the modality discriminator, which classifies whether the input feature representation is drawn from the visual or the textual modality. The discriminator consists of two fully-connected layers that reduce the embedding size to a scalar value which is used to predict the input modality. The discriminator is optimized according to the following GAN~\cite{goodfellow2014generative} loss function:
	\begin{equation}
	L_D = -\E_{V_i\sim V}[\log D\left(\phi(V_i)\right)] -\E_{T_i\sim T}[\log \left(1 - D\left(\tau(T_i)\right)\right)], 
	\end{equation}
	where \(V\) and \(T\) correspond to the image and text modalities respectively where samples are drawn and fed through the backbone architectures.  
	
	\begin{figure}[t]
		\centering
		\includegraphics[width=0.99\linewidth]{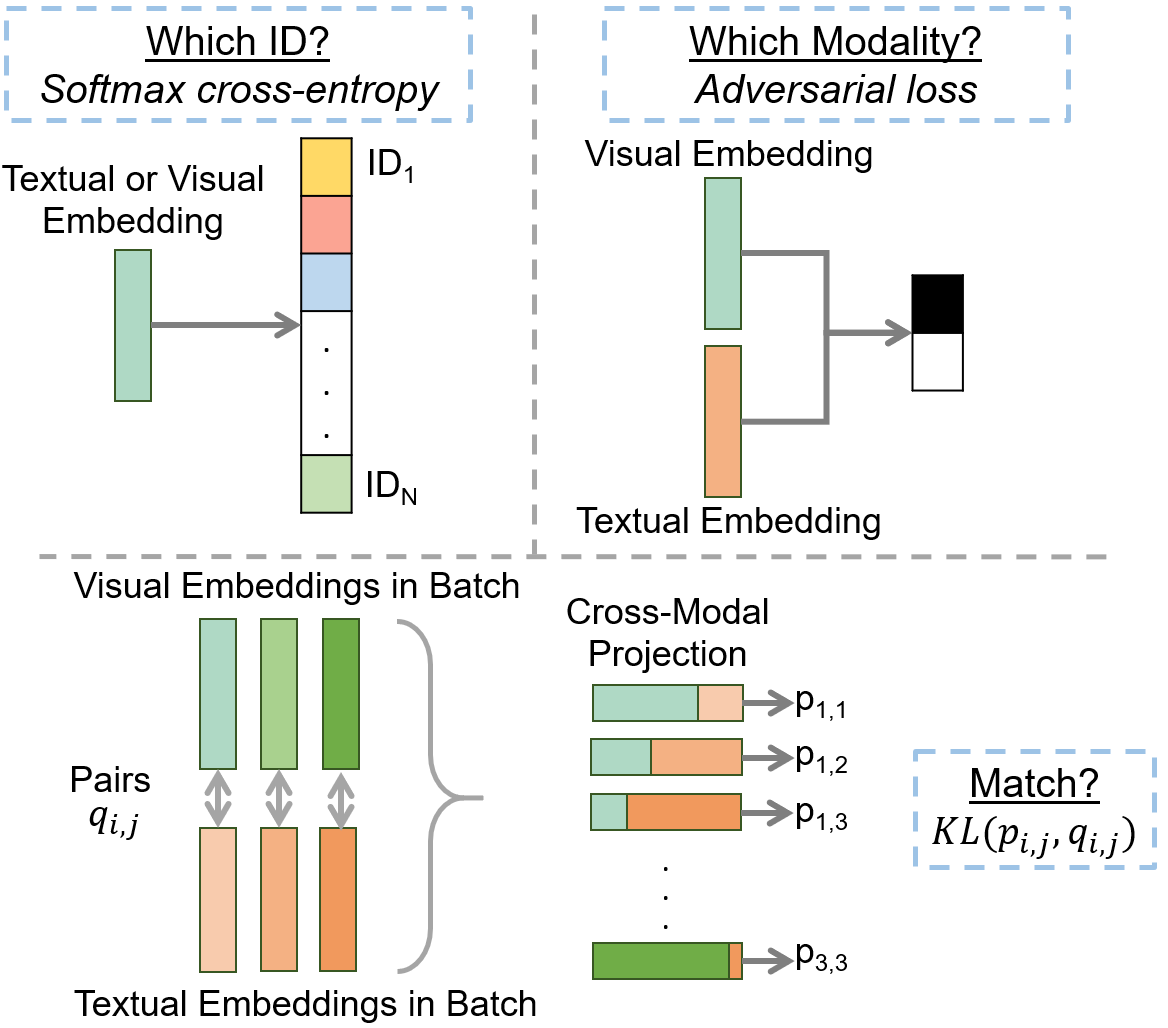}
		\caption{The three learning objectives. Top-left: We learn to classify each embedding based on the input ID. Top-right: We learn modality-invariant features using the discriminator. Bottom: We compute cross-modal projections between all samples in a batch in which samples from the same pair have a larger scalar projection, and learn to match using the predicted and the true matching probabilities.}
		\label{fig:losses}
		\vspace{-0.35cm}
	\end{figure}
	
	\subsection{Training and Testing Details}
	The loss function that is used to train TIMAM is the summation of two identification losses (\(L_I\)), the two cross-modal matching losses (\(L_M\)) and the adversarial loss of the discriminator (\(L_D\)): 
	\begin{equation}
	L = L_{I} + L_{M} + L_{D}
	\end{equation}
	An illustrative explanation of the three learning objectives is presented in Figure~\ref{fig:losses}. 
	We used stochastic gradient descent (SGD) with momentum equal to 0.9 to train the image and discriminator networks and the Adam optimizer~\cite{kingma2014adam} for the textual networks. The learning rate was set to \(2\times10^{-4}\) and was divided by ten when the loss plateaued at the validation set until \(2\times10^{-6}\). The batch-size was set to 64 and the weight decay to \(4\times10^{-4}\). The hidden dimension of the bidirectional LSTM was equal to 512 and the dimensionality of all feature vectors was set to 512. Finally, to properly balance the training between \(G^V\), \(G^T\), and \(D\), we followed several of the tricks discussed by Chintala \etal~\cite{chintala2016train}. For these balancing techniques along with the complete implementation details, a table with all notations as well as a detailed algorithm of our proposed approach, the interested reader is encouraged to refer to the supplementary material. 
	At testing time given a textual description as a probe, its textual features (\(\tau(T_i)\) extracted through the language backbone) and their distance between all image features (\(\phi(V_j)\) extracted from the image backbone) in the test set is computed using the cosine similarity:
	\begin{equation}
	s_{i,j} = \frac{\tau(T_i) \cdot \phi(V_j)}{||\tau(T_i)||\cdot||\phi(V_j)||}.
	\end{equation}
	The distances are then sorted and rank-1 through rank-10 results are reported. For image-to-text matching the same process is followed by using the image features as probe and retrieving the most relevant textual descriptions. 
	
	\begin{table}[t]
		\centering
		\caption{Text-to-image results (\%) on the CUHK-PEDES dataset. Results are ordered based on the rank-1 accuracy.}
		\small    
		\begin{tabular}{l rcc}
			\toprule
			\textbf{Method}  & {Rank-1} & {Rank-5} & {Rank-10}\\
			\midrule
			deeper LSTM Q+norm I~\cite{antol2015vqa} & 17.19 & - & 57.82 \\
			GNA-RNN~\cite{li2017person} & 19.05 & - & 53.64 \\
			IATV~\cite{li2017identity} & 25.94 & - & 60.48 \\
			PWM-ATH~\cite{chen2018improvingA} & 27.14 & 49.45 & 61.02 \\
			GLA~\cite{chen2018improving}& 43.58 & 66.93 & 76.26 \\
			Dual Path~\cite{zheng2017dual} & 44.40 & 66.26 & 75.07 \\
			CAN~\cite{jing2018cascade} & 45.52 & 67.12 & 76.98 \\
			CMPM + CMPC~\cite{zhang2018deep} & 49.37 & - & 79.27 \\
			\midrule
			\textbf{\methiccv} & \textbf{54.51} & \textbf{77.56} & \textbf{84.78}\\
			\bottomrule
		\end{tabular}%
		\vspace{-0.35cm}
		\label{tab:CUHK}%
	\end{table}%
	
	\section{Experiments}
	\noindent\textbf{Datasets}: To evaluate our method, four widely-used publicly available datasets were used and their evaluation protocols were strictly followed. We opted for these datasets in order to test TIMAM on a wide range of tasks ranging from pedestrians and flowers to objects and scenes. TIMAM was tested on (i) the CUHK-PEDES~\cite{li2017person} that contains images of pedestrians accompanied by two textual descriptions, (ii) the Flickr30K dataset~\cite{plummer2015flickr30k}, which contains a wide variety of images (humans, animals, objects, scenes) with five descriptions for each image, (iii) the Caltech-UCSD Birds (CUB)~\cite{reed2016learning} dataset that consists of images of birds with 10 descriptions for each image and finally, (iv) the Flowers~\cite{reed2016learning} dataset that consists of images of flowers originating for 102 categories with 10 descriptions for each image. 
	
	\begin{table*}[t]
		\centering
		\caption{Matching results on the Flickr30K dataset. The results are ordered based on their text-to-image rank-1 accuracy.}
		\small
		\begin{tabular}{lcccc|ccc}
			\toprule
			\multirow{2}{*}{\textbf{Method}} & \multirow{2}{*}{Image Backbone} & \multicolumn{3}{c}{Image-to-Text}
			& \multicolumn{3}{c}{Text-to-Image} \\
			\cmidrule(r){3-5} \cmidrule(r){6-8}
			& & Rank-1 & Rank-5 & Rank-10
			& Rank-1 & Rank-5 & Rank-10  \\
			\midrule
			DAN~\cite{nam2017dual} & VGG-19
			& {41.4} & {73.5} & {82.5}
			& {31.8} & {61.7} & {72.5} \\
			RRF-Net~\cite{liu2017learning} & ResNet-152
			& 47.6 & 77.4 & 87.1
			& 35.4 & 68.3 & 79.9  \\        
			CMPM +CMPC~\cite{zhang2018deep} & ResNet-152
			& {49.6} & {76.8} & {86.1}
			& {37.3} & {65.7} & {75.5}\\  
			DAN~\cite{nam2017dual} & ResNet-152
			& {55.0} & {81.8} & {89.0}
			& {39.4} & {69.2} & {79.1}\\
			NAR~\cite{liu2019neighbor} & ResNet-152
			& {55.1} & {80.3} & {89.6}
			& {39.4} & {68.8} & {79.9}\\
			VSE++~\cite{faghri2017vse} & ResNet-152
			& {52.9} & {80.5} & {87.2}
			& {39.6} & {70.1} & {79.5}\\
			SCO~\cite{huang2018learning} & ResNet-152
			& {55.5} & {\textbf{82.0}} & {89.3}
			& {41.1} & {70.5} & {80.1}\\
			GXN~\cite{gu2018look} & ResNet-152
			& {\textbf{56.8}} & {-} & {\textbf{89.6}}
			& {41.5} & {-} & {80.1}\\
			\midrule
			\textbf{\methiccv} & ResNet-152 & 53.1 & 78.8 & 87.6 & \textbf{42.6} & \textbf{71.6} & \textbf{81.9} \\
			\bottomrule
		\end{tabular}
		\label{tab:flickr}
		\vspace{-0.35cm}
	\end{table*}
	
	\noindent\textbf{Evaluation Metrics}: The evaluation metrics used in each dataset are adopted. Thus, for the CUHK-PEDES and Flickr30K datasets rank-1, rank-5, and rank-10 results are presented for each method. For the CUB and Flowers datasets, the \(AP@50\) metric is utilized for text-to-image retrieval and rank-1 for image-to-text matching. Given a query textual class, the algorithm first computes the percentage of top-50 retrieved images whose identity matches that of the textual query class. The average matching percentage of all test classes is denoted as \(AP@50\). Finally, note that in each dataset TIMAM is evaluated against the eight best-performing methods. The complete results against all methods tested in each dataset as well as all details regarding the datasets (\eg, train/val/test splits, pre-processing) are included in the supplementary material.

	\subsection{Quantitative Results}
	\noindent\textbf{CUHK-PEDES Dataset}: We evaluate our approach against the eight best-performing methods that have been tested on the CUHK-PEDES dataset and present text-to-image matching results in Table~\ref{tab:CUHK}. Some key methods that have been evaluated on this dataset include (i) IATV~\cite{li2017identity} which learns discriminative features using two attention modules working on both modalities at different levels but it is not end-to-end; (ii) GLA~\cite{chen2018improving} which identifies local textual phrases and aims to find the corresponding image regions using an attention mechanism; (iii) and CMPM~\cite{zhang2018deep} in which two projection losses are proposed to learn features for text-to-image matching. TIMAM outperforms all previous works by a large margin. We observe an absolute improvement of more than 5\% in terms of rank-1 over the previous best performing method~\cite{zhang2018deep} which originates from learning better feature representations through the identification and cross-modal matching losses as well as the proposed adversarial learning framework. 
	
	\noindent\textbf{CUB and Flowers Datasets}: We test TIMAM against all eight methods evaluated on these datasets and present our matching results in Table~\ref{tab:cubfl}. Our method achieves state-of-the-art results in both image-to-text and text-to-image matching in both datasets. We observe performance increases of 2.2\% and 3.4\% in terms of rank-1 accuracy as well as 3.6\% and 2.4\% in terms of AP@50. 
	
	\noindent\textbf{Flickr30K Dataset}: In Table~\ref{tab:flickr}, we report cross-modal retrieval results on the Flickr30K dataset against the top-8 best-performing methods. Similar to the best-performing methods, and only in this dataset, a ResNet-152 is employed to allow for a fair comparison. TIMAM surpasses all methods by a large margin in text-to-image matching but demonstrates inferior performance compared to GXN~\cite{gu2018look} in image-to-text matching. Most of the best-performing image-to-text matching methods employ multi-step attention blocks and thus, are able to learn ``where to look'' in an image which results in better image features. Unlike the rest of the datasets that contain a single primary object (\ie, flowers/birds/pedestrians only), Flickr30K contains a wide range of primary components. This image variance coupled with the relatively small number of training images make cross-modal matching a challenging task. While our approach achieves state-of-the-art results in the text-to image matching task and is capable of learning correct associations between images and descriptions, there is still room for further improvements by future research. 
	
	\setlength{\tabcolsep}{.1cm}
	\begin{table}[t]
		\centering
		\caption{Cross-modal matching results on the CUB and Flowers datasets. The results are ordered based on the text-to-image AP@50 performance.}
		\label{tab:cubfl}
		\small
		\begin{tabular}[t]{lcc|cc}
			\toprule
			\multirow{3}{*}{\textbf{Method}} & \multicolumn{2}{c}{\textbf{CUB}} & \multicolumn{2}{c}{\textbf{Flowers}} \\
			& Img2Txt & Txt2Img & Img2Txt & Txt2Img\\
			\cmidrule{2-2}\cmidrule{3-3}\cmidrule{4-4}\cmidrule{5-5}
			& Rank-1 & AP@50 & Rank-1 & AP@50\\
			\midrule
			Word2Vec~\cite{mikolov2013distributed} & 38.6 & 33.5 & 54.2 & 52.1 \\
			GMM+HGLMM~\cite{klein2015associating} & 36.5 & 35.6 & 54.8 & 52.8 \\
			Word CNN~\cite{reed2016learning} & 51.0 & 43.3  & 60.7 & 56.3\\
			Word CNN-RNN~\cite{reed2016learning} & 56.8 & 48.7 & 65.6 & 59.6 \\
			Attributes~\cite{akata2015evaluation} & 50.4 & 50.0 & - & - \\
			Triplet Loss~\cite{li2017identity} & 52.5 & 52.4  & 64.3 & 64.9 \\
			IATV~\cite{li2017identity} & 61.5 & 57.6 & 68.9 & 69.7 \\
			CMPM+CMPC~\cite{zhang2018deep} & 64.3 & 67.9  & 68.4 & 70.1 \\
			\midrule
			\textbf{\methiccv} & \textbf{67.7} & \textbf{70.3} & \textbf{70.6} & \textbf{73.7}\\
			\bottomrule
		\end{tabular}
		\vspace{-0.3cm}
	\end{table}
	
	\subsection{Ablation Studies}\label{ssec:abl}
	\noindent\textbf{Impact of Proposed Components}: In our first ablation study (Table~\ref{tab:cuhk_abl}), we assess how each proposed component of TIMAM contributes to the final text-to-image matching performance on the CUHK-PEDES dataset. We observe that the identification (\(\mathcal{L}_{I}\)) and cross-modal projection (\(\mathcal{L}_{M}\)) losses result in a rank-1 accuracy of 49.85\% when used together and considerably less when used individually. By introducing BERT, better word embeddings can be learned that increase the accuracy to 52.97\%. Finally, when the proposed adversarial representation learning paradigm (\arl) is used, additional improvements are observed, regardless of whether BERT is used or not. We observe relative improvements of 2.9\% and 3\% with and without BERT respectively, which demonstrates that ARL helps the network learn modality-invariant representations that can successfully be used at deployment-time to perform cross-modal matching. Similar results are obtained in the Flickr30K dataset in which ARL improved the rank-1 matching performance from 51.2\% to 53.1\% and from 41.0\% to 42.6\% in image-to-text and text-to-image respectively. 
	\setlength{\tabcolsep}{0.25cm}
	\begin{table}[t]
		\centering
		\caption{Ablation studies on the CUHK-PEDES dataset to investigate the additions in terms of rank-1 and rank-10 accuracy for the identification (\(\mathcal{L}_{I}\)) and cross-modal projection (\(\mathcal{L}_{M}\)) losses, the addition of BERT as a backbone architecture for language modeling, and the adversarial representation learning paradigm.}
		\label{tab:cuhk_abl}
		\begin{tabular}[t]{cccc|cc}
			\toprule
			\(\mathcal{L}_{I}\) & \(\mathcal{L}_{M}\) & BERT & \arl & Rank-1 & Rank-10 \\
			\midrule
			\checkmark & & & & 40.1 & 70.1 \\
			& \checkmark & & & 44.9 & 77.7 \\
			\checkmark & \checkmark & & &  49.8 &  81.5 \\
			\checkmark & \checkmark & & \checkmark & 51.3 & 82.4 \\    
			\checkmark & \checkmark & \checkmark & &  52.9 &  83.5 \\
			\checkmark & \checkmark & \checkmark & \checkmark &  54.5 &  84.8 \\
			\bottomrule
		\end{tabular}
		\vspace{-0.35cm}
	\end{table}
	
	\noindent\textbf{Impact Backbone Depth}: In the second ablation study, we investigate to what extent the depth of the backbone networks affects the final performance. Similar to previous well-performing methods~\cite{chen2018improving, li2017identity, zhang2018deep}, a fully-connected layer was used to learn the word embeddings (denoted by FC-Embed.) and its impact was compared with the deep language model of BERT. For the image modality, two different ResNet backbones were employed while the rest of our proposed methodology remained the same. Rank-1 matching results in both directions are reported on the Flickr30K dataset in Table~\ref{tab:flickr_abl}. Introducing a language model yields significant improvements (4.8\% and 4.7\%) regardless of the image backbone. In addition, increasing the image backbone depth results in smaller text-to-image matching improvements of approximately 2\%.

\begin{figure*}[t]
		\centering
		\includegraphics[width=0.95\linewidth]{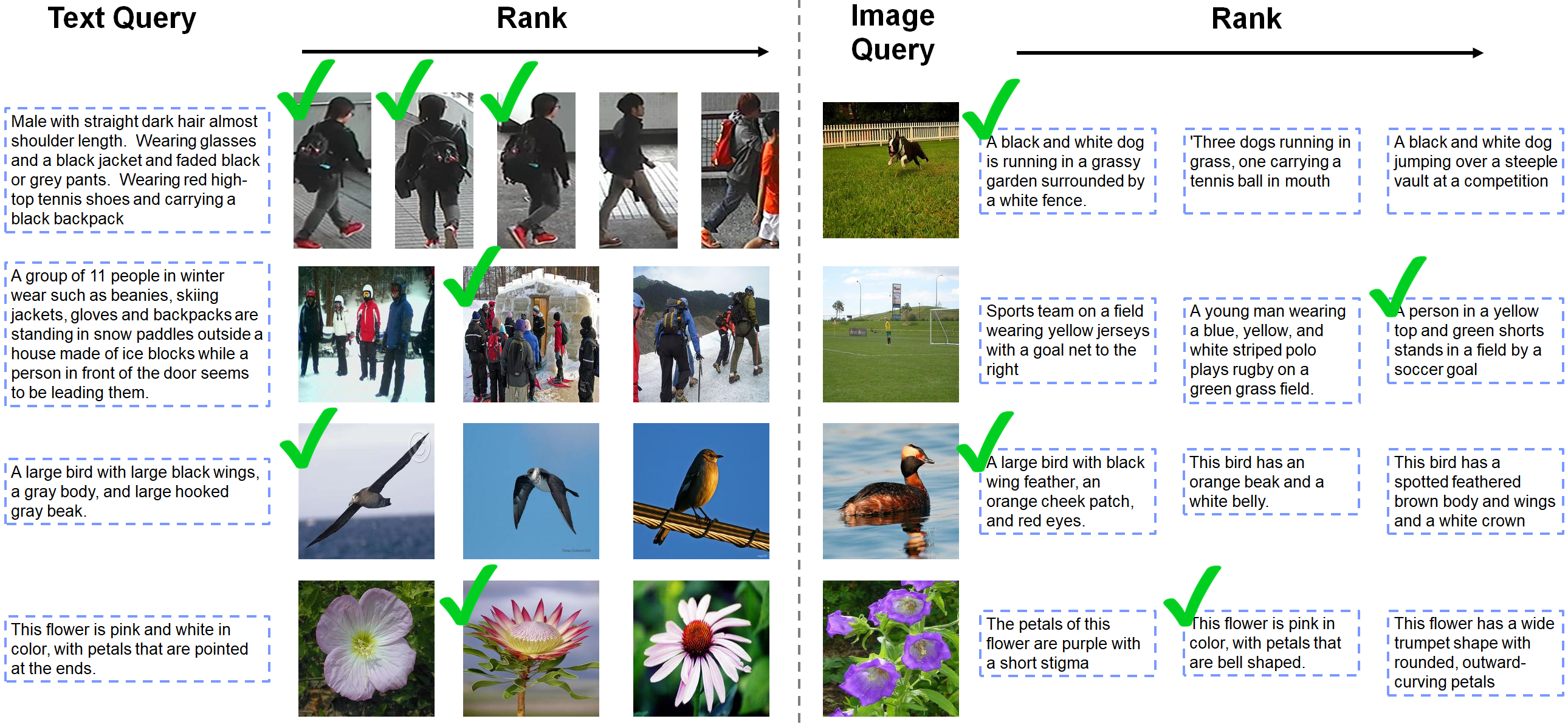}
		\caption{Qualitative results on all datasets we tested our method. Given a textual/visual description as a query, we retrieve the most relevant images/descriptions ranked from left to right. Successful retrieval is achieved in cases with poor lighting, under different poses, and with different visual attributes. Even in failure cases, we observe that the retrieved results are still very relevant (\eg, in the top left example the \(4^{th}\) and \(5^{th}\) image matches contain individuals with gray pants and a backpack).}
		\vspace{-0.3cm}
		\label{fig:qual1f}
	\end{figure*}

\begin{figure}[t]
		\centering
		\includegraphics[width=0.97\linewidth]{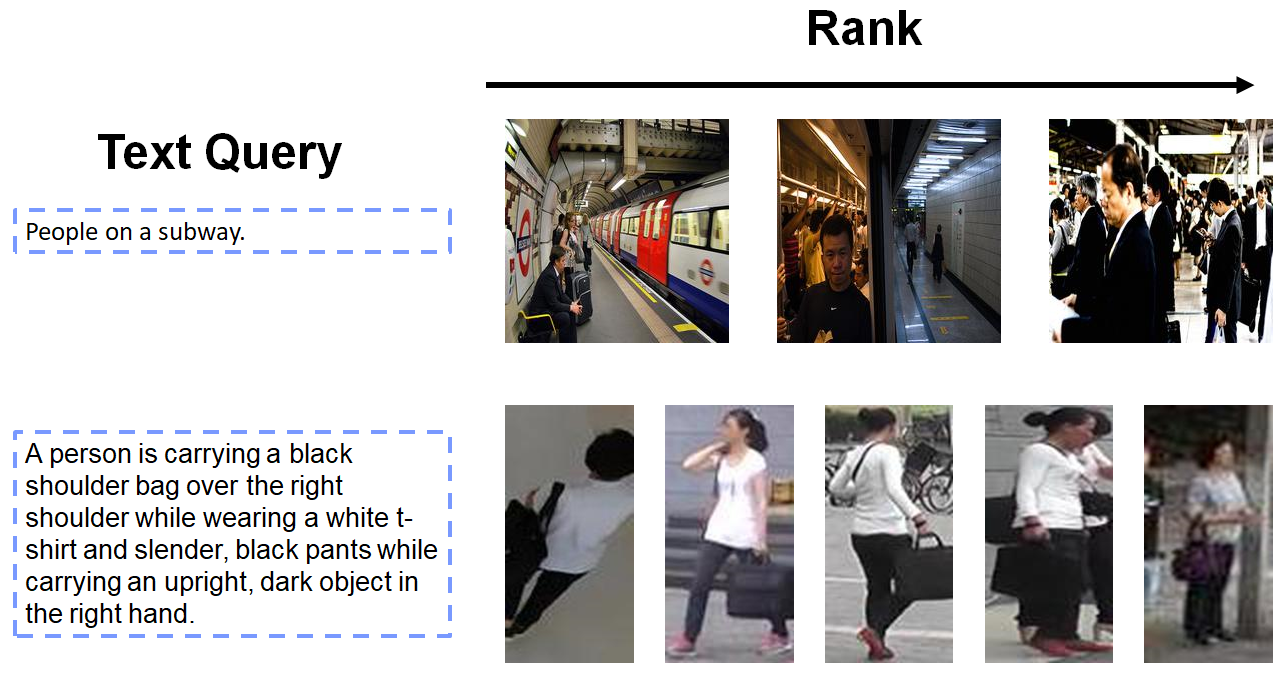}
		\caption{Two failure cases of the proposed approach. While neither of the retrieved results match the true text ID, they are still very relevant to the textual query.}
		\label{fig:failures}
        \vspace{-0.2cm}
	\end{figure}

	\noindent\textbf{Qualitative Results}: Figure~\ref{fig:qual1f} depicts cross-modal retrieval results for all four datasets. We observe that TIMAM is capable of learning cloth and accessory-related correspondences as it can accurately retrieve images of people carrying bags with the correct set of clothing. The proposed approach retrieves consistent images given a textual query (\eg, group of people in snow) as well as similar textual descriptions, given an image query (\eg, all three descriptions describe dogs or soccer players in the first and second row on the right). Finally, some interesting observations can be made from the failure cases. While all retrieved images in Figure~\ref{fig:failures} have a different ID than the true label of the textual description, TIMAM can still retrieve images that match the textual input. For example, all images in the second row contain a female with a white t-shirt, black pants, carrying a bag. 
	
	\setlength{\tabcolsep}{.1cm}
	\begin{table}[t]
		\centering
		\caption{Ablation studies on the Flickr30K dataset to assess the impact of the depth of different backbone architectures.}
		\label{tab:flickr_abl}
		\small
		\resizebox{\columnwidth}{!}{
			\begin{tabular}[t]{cc|cc||cc}
				\toprule
				\multicolumn{2}{c}{Image Backbone} & \multicolumn{2}{c}{Text Backbone} & Img2Txt & Txt2Img \\
				\cmidrule(r){1-2} \cmidrule(r){3-4} \cmidrule(r){5-5} \cmidrule(r){6-6}
				ResNet-101 & ResNet-152 & FC-Emb. & BERT & Rank-1 & Rank-1\\
				\midrule
				\checkmark & & \checkmark & & 47.9 & 35.8 \\
				\checkmark & & &\checkmark  & 52.0 & 40.6 \\
				& \checkmark & \checkmark & & 50.1 & 37.9   \\
				& \checkmark & & \checkmark & 53.1 & 42.6 \\    
				\bottomrule
			\end{tabular}
		}
		\vspace{-0.3cm}
	\end{table}
	
	\subsection{Discussion of Alternatives}
	\noindent\textbf{Loss Functions}: The wide variety of loss functions available in the literature that could potentially be applicable to our problem, begs the question of why did we choose these specific identification and matching losses instead of other alternatives? Our first goal was to refrain from using losses that sample triplets or quadruplets within each batch~\cite{Chen_2017_CVPR, dong2017class, huang2016learning, schroff2015facenet}. The reason is that such losses introduce a computational overhead during training~\cite{schroff2015facenet, taha2019defense} and additional hard-mining schemes~\cite{dong2017class} would have to be incorporated to ensure that hard negatives are provided to such loss functions. Second, we relied on the experimental investigation of Zhang and Lu~\cite{zhang2018deep} that showed the KL-based loss described in Eq.~(\ref{eq:cmpm}) achieves superior matching performance among other alternatives~\cite{liu2017learning, sohn2016improved, ustinova2016learning} and is robust to small or large batch sizes. Finally, we demonstrate through the ablation studies described in Section~\ref{ssec:abl}, that the identification and matching losses we introduced result in substantial improvements in terms of rank-1 accuracy.
	
	\noindent\textbf{Text Augmentation}: Aiming to introduce some noise to the textual input as a form of data-augmentation that could potentially improve the performance we experimented with the conditional augmentation technique of Zhang \etal~\cite{zhang2017stackgan}. Conditional augmentation resembles the reparametrization trick used in variational autoencoders, as it maps the text embedding to a mean and a variance feature vector which are then added together using some noise sampled from \(\mathcal{N}(0, I)\). In that way, more training pairs are generated given a small number of image-text pairs, and the method is robust to small perturbations in the conditioning manifold. However, when we experimented with this technique on the CUHK-PEDES dataset, we observed consistently worse results compared to the original approach. We thus believe that conditional augmentation is not suitable for text-to-image matching as not only two additional embeddings need to be learned (\ie, more parameters), but also the noise that is introduced ends up obfuscating the model instead of augmenting its learning and generalization capabilities. 
	
	\noindent\textbf{Text-to-Image Reconstruction}: Aiming to learn textual features capable of retrieving the most relevant images, we experimented with text-to-image reconstruction as an additional learning objective trained in an end-to-end setup. The textual embedding was fed to a decoder comprising upsampling and convolution layers that reconstructed the corresponding input image at different scales. While this method demonstrated good reconstruction results in the Flowers and CUB datasets, which is in line with existing work~\cite{qiao2019mirrorgan, xu2018attngan, zhang2017stackgan}, this was not the case for the CUHK-PEDES and Flickr30K datasets. In the latter, the reconstructions were very blurry (\eg, only the general human shape was visible for pedestrians) which is explained by the large variance of the input images and thus, could not help us learn better features. While birds and flowers follow a very similar image pattern, this is not the case with images in Flickr30K which contain a wide range of objects or humans performing different actions from different viewpoints. 
	
	\section{Conclusion}
	Learning discriminative representations for cross-modal matching has significant challenges such as the large variance of the linguistic input and the difficulty of measuring the distance between the multi-modal features. To address these challenges, we introduced TIMAM: a text-image matching approach that employs an adversarial discriminator that aims to identify whether the input is originating from the visual or textual modality. When the discriminator is jointly trained with identification and cross-modal matching objectives, it results in discriminative modality-invariant embeddings. In addition, we observed that a deep language model can boost the cross-modal matching capabilities since better textual embeddings are learned. 
	We demonstrated through extensive experiments, that (i) adversarial learning is well-suited for text-image matching, and (ii) a deep language model can successfully be utilized in cross-modal matching applications. State-of-the-art results were obtained in four publicly available datasets all of which are widely used in this domain. To facilitate further investigation by future research, we performed ablation studies, discussed alternative approaches that were explored, and presented qualitative results that provide an insight into the performance of our approach. 
	\linebreak
	\section*{Acknowledgments}
	 This work has been funded in part by the UH Hugh Roy and Lillie Cranz Cullen Endowment Fund. All statements of fact, opinion or conclusions contained herein are those of the authors and should not be construed as representing the official views or policies of the sponsors.
	\clearpage
	{\small
		\bibliographystyle{ieeefullname}
		\bibliography{Refs}
	}
	\clearpage
	\section*{Supplementary Material}
	\begin{table*}
		\centering
		\caption{Notation used throughout our paper}
		\small
		\begin{tabular}{cl}
			\toprule
			\textbf{Notation Sign} & \textbf{Description} \\
			\midrule
			\(V\) & The visual (\ie, image) modality\\
			\(T\) & The textual modality\\
			\(V_i\) & A sample from the visual modality\\
			\(T_i\) & A sample from the textual modality\\
			\(Y_i\) & The ID/Category label of the pair\\
			\midrule
			\(\phi(\cdot)\) & The feature extractor at the image modality (\ie, ResNet-101) \\
			\(\tau(\cdot)\) & The feature extractor at the textual modality (\ie, BERT, the LSTM and the FC- layer)\\
			\(\bar{\tau}(\cdot)\) & Normalized textual features\\
			\midrule
			\(G^V, G^T\) & Generators from the visual and textual modality (\ie, \(\phi()\) and  \(\tau()\)) \\
			\(D\) & Cross-modal discriminator\\
			\midrule
			\((W_i, b_i)\) & Weights and bias of the last FC-layer that produces the embedding\\
			\(B\) & Batch size\\
			\midrule
			\(p_{i,j}\) & Probability of matching each visual embedding to each normalized textual embedding in the batch\\
			\(q_{i,j}\) & True matching probability for each pair in the batch\\
			\(s_{i,j}\) & Cosine similarity between  \(i^{th}\) probe and \(j^{th}\) gallery sample \\
			\midrule
			\(L_{I}^{V}\) & Norm-softmax cross entropy loss used for identification for the visual embedding\\
			\(L_{I}^T\) & Norm-softmax cross entropy loss used for identification for the textual embedding\\
			\(L_{I}\) & Summation of the two identification losses from both modalities\\
			\(L_{M}^V\) & KL-divergence loss used for cross-modal \((V->T)\) projection matching \\
			\(L_{M}^T\) & KL-divergence loss used for cross-modal \((T->V)\) projection matching \\
			\(L_M\) & Summation of the two cross-modal projection losses from both modalities\\
			\(L_D\) & Adversarial loss of the discriminator\\
			\(L\) & Loss used to train our network: summation of individual sub-losses\\
			\bottomrule
		\end{tabular}
		\label{tab:notation}
		\vspace{0.2cm}
	\end{table*}

	\section*{Discussion on Novelty}
	What is novel in TIMAM? What has been done and what is new in the proposed approach? Since all these are valid questions that the interested reader might have, we aim to provide the reader with an in-depth understanding of the contributions of our work and how these result in advantages over previous work.
	
	Our method is novel in that we take a different approach in learning the embeddings from both modalities. Specifically, we leverage an adversarial discriminator, which helps TIMAM to learn modality-invariant discriminative representations. It is a very effective addition that can easily be applied to other cross-modal matching applications (\eg, audio-visual retrieval).   
	Our ablation studies, in two very challenging datasets, showed improvements of \mytilde3\% on the CUHK-PEDES dataset and \mytilde1.8\% on the Flickr30K dataset, when the adversarial discriminator is used. These results demonstrate that adversarial learning is well-suited for cross-modal matching. 
	
	The second contribution of this work is that we demonstrated that a pre-trained language model can successfully be applied (with some fine-tuning) to computer vision applications such as text-to-image matching. Our results demonstrated that we can improve our feature representations when better word embeddings are learned in this manner. In summary, the advantages of TIMAM over prior work can be described as follows: 
	\begin{itemize}[noitemsep]
		\item[--] TIMAM improves upon CMPM~\cite{zhang2018deep} (previous best performing method on the CUHK-PEDES, Flowers, and Birds datasets) by employing a domain discriminator, which results into more discriminative representations. Unlike CMPM, we do not perform cross-modal projections in the classification loss since we observed that their contribution is insignificant. Instead we employ identification losses for the visual and textual features and present their impact in the first ablation study. 
		\item[--] TIMAM improves upon the previous best performing method on the Flickr30K dataset by learning better textual embeddings using the fine-tuning capabilities of BERT (as well as employing the adversarial representation learning framework). 
		\item[--] TIMAM is very easy to reproduce, which is not the case with prior work that requires complex attention mechanisms at both modalities~\cite{nam2017dual} or text reconstruction objectives~\cite{chen2018improving}. To obtain our results one can simply fetch an image backbone and a deep language model and then follow the steps described in Alg.~\ref{alg1}.  
	\end{itemize}
	
	\setlength{\textfloatsep}{12pt}
	\begin{algorithm}[t]
		\SetKwInOut{Input}{Input}
		\SetKwInOut{Output}{Output}
		\Input{Batch (\(B\)) of image-text pairs \((V_i, T_i)\) with their label \(Y_i\), pre-trained ResNet-101 weights, pre-trained BERT weights}
		\(\phi(V_i)\leftarrow\) extract visual embedding by feeding \(V_i\) to image backbone\\
		\(\tau(T_i)\leftarrow\) extract textual embedding by feeding \(T_i\) to text backbone and then to the LSTM\\
		\(L_{I}^{V}\leftarrow\) compute identification loss for the images using \((V_i, Y_i)\). Similarly compute \(L_{I}^{T}\) for the text.\\
		\(p_{i,j}\leftarrow\) compute the probability of matching \(\phi(V_i)\) to \(\bar{\tau}(T_j)\) \\
		\(q_{i,j}\leftarrow\) compute the true matching probability using \(Y_i\) as well as the rest of true labels in \(B\)) \\
		\(L_M^V\leftarrow\) compute cross-modal projection matching loss as the KL divergence from \(q_i\) to \(p_i\)\\
		Repeat steps 4-6 for the text modality to compute \(L_M^T\) by normalizing \(\phi(V_i)\) instead of \(\tau(T_i)\) \\
		\(L_D\leftarrow\) compute adversarial loss by passing \(\phi(V_i)\) and \(\tau(T_i)\) through the discriminator \\
		Update network parameters using: \(L=L_D + L_{I}^{V} + L_{I}^{T} + L_M^V + L_M^T\)\\
		\Output{Network weights}
		\caption{Training Procedure of TIMAM}
		\label{alg1}
	\end{algorithm}

	\section*{Implementation Details}
	\noindent\textbf{Datasets}: The first dataset used was the CUHK-PEDES~\cite{li2017person} which consists of 40,206 images of individuals of 13,003 identities, and each image is described by two textual descriptions. The dataset is split into 11,003/1,000/1,000 identities for the training/validation/testing sets with 34,054, 3,078 and 3,074 images respectively, in each subset. The second dataset was the Flickr30K~\cite{plummer2015flickr30k} which contains 31,783 images with five text descriptions each. The data split introduced in the work of Karpathy and Fei-Fei~\cite{karpathy2015deep} is adopted which results in 29,783/1,000/1,000 images for training validation and testing respectively. The third dataset was the Caltech-UCSD Birds (CUB)~\cite{reed2016learning}, which comprises 11,788 bird images from 200 different categories. Each image is labeled with 10 descriptions and the dataset is split into 100 training, 50 validation and 50 test categories. Finally, the Oxford102 Flowers (Flowers)~\cite{reed2016learning} dataset was used, which consists of 8,189 flower images of 102 different categories. Each image is accompanied by 10 descriptions and the dataset is split into 62 training, 20 validation, and 20 test categories. 
	
	\noindent\textbf{Data Pre-processing}: For the CUHK-PEDES dataset all images were resized to \(128\times256\) since pedestrians walking are usually rectangular. For the rest of the datasets, all images were resized to \(224\times224\). For the textual input, basic word tokenization was performed by mapping each word to the vocabulary accompanying the base BERT model pre-trained on the uncased book corpus and English Wikipedia datasets.\footnote{The pre-trained model we used is available at the Gluon-NLP website: \url{https://gluon-nlp.mxnet.io/model_zoo/bert/index.html}} For the CUHK-PEDES dataset the maximum length of the sentences was set to 50 words (following the pre-processing steps of Li \etal~\cite{li2017person}) whereas for the rest of the datasets it was set to 30 words (following the pre-processing steps of Zhang and Lu~\cite{zhang2018deep} for Flickr30K and Reed \etal~\cite{reed2016learning} for the CUB and Flowers datasets). Thus, sentences shorter than the maximum length were zero-padded, whereas those longer than the threshold were trimmed. 
	
	\noindent\textbf{Data Augmentation}: During data augmentation images were upscaled to \(\times1.25\) the original size in both dimensions and random crops of the original dimensions were extracted and fed to the model. In addition, data shuffling, random horizontal flips with 50\% probability and color jittering were employed. 
	
	\noindent\textbf{Architecture Details}: We used the pre-trained models of ResNet-101 and BERT available online for the backbone architectures of the two modalities while the rest of the layers were initialized with Xavier initialization. 
	\begin{itemize}
		\item \textbf{Image domain}: Our backbone architecture on the visual domain is a ResNet-101 that extracts feature representations of  dimensionality \(7\times7\times2,048\) (for an input image with dimensions \(224\times224\times3\)). These representations are then fed to a fully-connected layer after performing global-average pooling to extract the image embedding of size equal to 512. 
		\item \textbf{Text  domain}: For the textual domain, each tokenized input sentence of length is fed to the deep language model which extracts a 768-D vector for each word. The sequence of word embeddings is then fed to a bidirectional LSTM with 512 hidden dimensions and its output is then projected to a fully-connected layer which outputs the text embedding of size equal to 512.
		\item \textbf{Discriminator}: We opted for a simple discriminator comprising two fully-connected layers [FC(256)-BN-LReLU(0.2)-FC(1)] that reduce the embedding size to a scalar value which is used to predict the input domain. 
	\end{itemize}  
	
	\begin{figure*}[t]
		\centering
		\includegraphics[width=\linewidth]{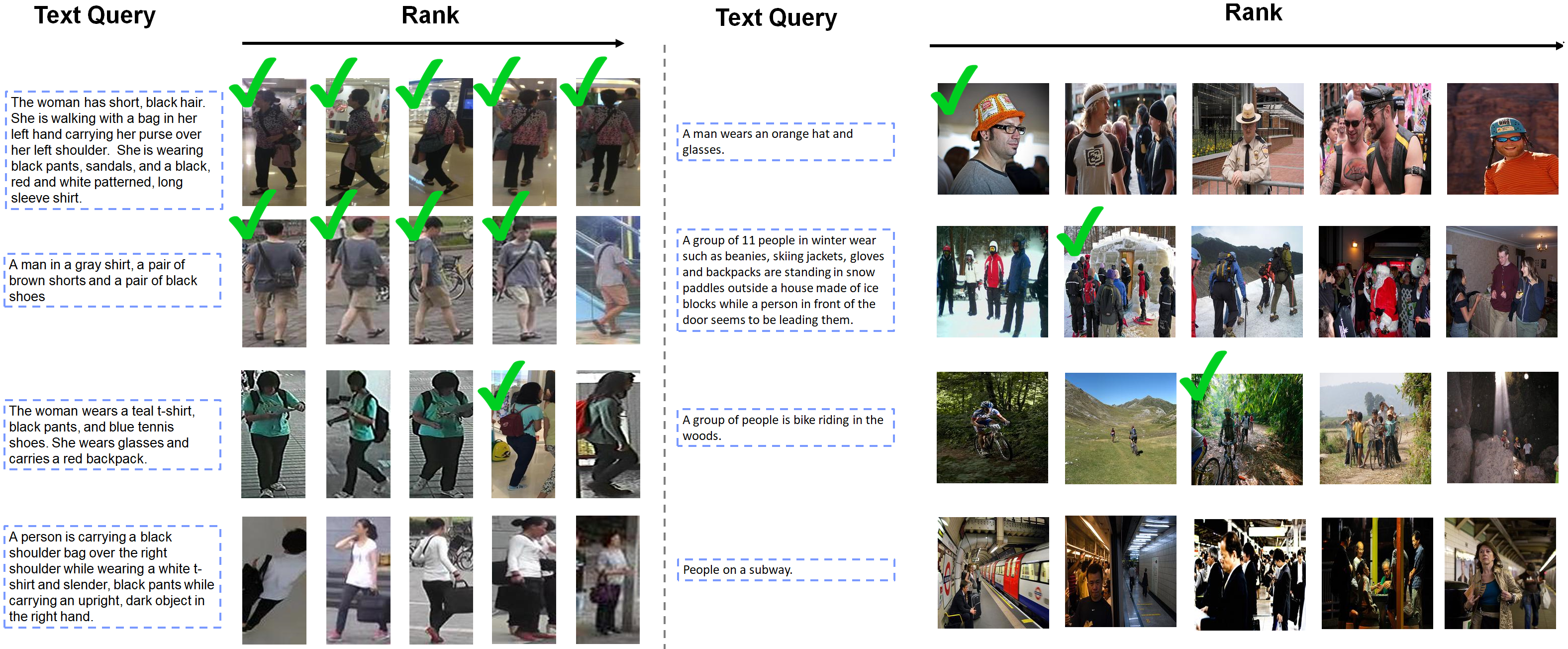}
		\caption{Additional qualitative text-to-image retrieval results on the CUHK-PEDES (left) and Flickr30K (right) datasets.}
		\label{fig:qual2f}
	\end{figure*}
	
	\noindent\textbf{Training Details}:  We present all the notation used throughout our work in Table~\ref{tab:notation} for easier reference. We used MXNet/Gluon as our deep learning framework and a single NVIDIA GeForce GTX 1080 Ti GPU. We used stochastic gradient descent (SGD) with momentum equal to 0.9 to train the image and discriminator networks and the Adam optimizer~\cite{kingma2014adam} for the textual networks. The learning rate was set to \(2\times10^{-4}\) and was divided by ten when the loss plateaued at the validation set until \(2\times10^{-6}\). The batch-size was set to 64 and the weight decay to \(4\times10^{-4}\). The deep language model was initially frozen and the rest of the parameters were updated until convergence. After this step, we unfroze its weights and the whole network was fine-tuned with a learning rate equal to \(2\times10^{-6}\) for 30 epochs. Successfully, training the discriminator required maintaining an adequate balance between the two feature generators and the discriminator. To accomplish that, we relied on several of the tricks presented by Chintala \etal~\cite{chintala2016train} on how to train a GAN: (i) different mini-batches were constructed for the features of each domain, (ii) labels were smoothed by replacing each positive label (visual domain) with a random number in [0.8, 1.2], and each label equal to zero (textual domain) with a random number in [0, 0.3], and (iii) labels were flipped with 20\% probability to introduce some noise. 
	
	\section*{Extended Quantitative Results}
	Due to space constraints in the main paper, we provided quantitative results that contained the 8 best-performing methods in each dataset. In Tables~\ref{tab:CUHK1} and \ref{tab:flickr1}, we present complete results against all approaches test in the CUHK-PEDES and Flickr30K datasets. TIMAM surpasses all methods in text-to-image matching but demonstrates inferior performance compared to GXN~\cite{gu2018look} in image-to-text matching.       
	
	\begin{table}[t]
		\centering
		\caption{Text-to-image results on the CUHK-PEDES dataset. Results are ordered based on the rank-1 accuracy.}
		\small    
		\begin{tabular}{l rcc}
			\toprule
			\textbf{Method}  & {Rank-1} & {Rank-5} & {Rank-10}\\
			\midrule
			iBOWIMG~\cite{zhou2015simple} & 8.00 & - & 30.56 \\
			Word CNN-RNN~\cite{reed2016learning} & 10.48 & - & 36.66 \\
			Neural Talk~\cite{vinyals2015show} & 13.66 & - & 41.72 \\
			GMM+HGLMM~\cite{klein2015associating} & 15.03 & - & 42.47 \\
			deeper LSTM Q+norm I~\cite{antol2015vqa} & 17.19 & - & 57.82 \\
			GNA-RNN~\cite{li2017person} & 19.05 & - & 53.64 \\
			IATV~\cite{li2017identity} & 25.94 & - & 60.48 \\
			PWM-ATH~\cite{chen2018improvingA} & 27.14 & 49.45 & 61.02 \\
			GLA~\cite{chen2018improving}& 43.58 & 66.93 & 76.26 \\
			Dual Path~\cite{zheng2017dual} & 44.40 & 66.26 & 75.07 \\
			CAN~\cite{jing2018cascade} & 45.52 & 67.12 & 76.98 \\
			CMPM + CMPC~\cite{zhang2018deep} & 49.37 & - & 79.27 \\
			\midrule
			\textbf{\methiccv} & \textbf{54.51} & \textbf{77.56} & \textbf{84.78}\\
			\bottomrule
		\end{tabular}%
		\label{tab:CUHK1}%
		\vspace{-0.35cm}
	\end{table}
	
	\section*{Extended Qualitative Results}
	In Figure~\ref{fig:qual2f} we present additional qualitative text-to-image matching results on the CUHK-PEDES and Flickr30K datasets. We observe that TIMAM is capable of learning visual attributes related to soft-biometrics (\eg, sex of the individual), clothing (gray t-shirts on second row to the left or teal t-shirts on the third row) as well as objects such as hats (first row to the right) and backpacks (third and fourth rows to the left). To our surprise, we can effectively learn to match descriptions and images of scenes/actions (biking in the woods, or people on a subway) while having only a handful of such examples in the whole dataset. 
	
	\begin{table*}
		\centering
		\caption{Matching results on the Flickr30K dataset. The results are ordered based on their text-to-image rank-1 accuracy.}
		\small
		\begin{tabular}{lcccc|ccc}
			\toprule
			\multirow{2}{*}{\textbf{Method}} & \multirow{2}{*}{Image Backbone} & \multicolumn{3}{c}{Image-to-Text}
			& \multicolumn{3}{c}{Text-to-Image} \\
			\cmidrule(r){3-5} \cmidrule(r){6-8}
			& & Rank-1 & Rank-5 & Rank-10
			& Rank-1 & Rank-5 & Rank-10  \\
			\midrule
			DVSA~\cite{karpathy2015deep} & RCNN
			& 22.2 & 48.2 & 61.4
			& 15.2 & 37.7 & 50.5  \\
			{m-RNN-VGG~\cite{mao2014deep}} & VGG-19
			& {35.4} & {63.8} & {73.7}
			& {22.8} & {50.7} & {63.1} \\
			{HGLMM FV~\cite{plummer2015flickr30k}} & VGG-19
			& {36.5} & {62.2} & {73.3}
			& {24.7} & {53.4} & {66.8}  \\
			VQA-A~\cite{lin2016leveraging} & VGG-19
			& 33.9 & 62.5 & 74.5
			& 24.9 & 52.6 & 64.8  \\
			{GMM+HGLMM~\cite{klein2015associating}} & VGG-19
			& {35.0} & {62.0} & {73.8} 
			& {25.0} & {52.7} & {66.0}  \\
			{m-CNN~\cite{ma2015multimodal}} & VGG-19
			& {33.6} & {64.1} & {74.9} 
			& {26.2} & {56.3} & {69.6} \\
			DCCA~\cite{yan2015deep} & AlexNet
			& {27.9} & {56.9} & {68.2}
			& {26.8} & {52.9} & {66.9}  \\
			{DSPE~\cite{wang2016learning}} & VGG-19
			& {40.3} & {68.9} & {79.9} 
			& {29.7} & {60.1} & {72.1}  \\
			sm-LSTM~\cite{huang2017instance} & VGG-19
			& 42.5 & 71.9 & 81.5
			& 30.2 & 60.4 & 72.3  \\
			DAN~\cite{nam2017dual} & VGG-19
			& {41.4} & {73.5} & {82.5}
			& {31.8} & {61.7} & {72.5} \\
			RRF-Net~\cite{liu2017learning} & ResNet-152
			& 47.6 & 77.4 & 87.1
			& 35.4 & 68.3 & 79.9  \\        
			CMPM +CMPC~\cite{zhang2018deep} & ResNet-152
			& {49.6} & {76.8} & {86.1}
			& {37.3} & {65.7} & {75.5}\\  
			DAN~\cite{nam2017dual} & ResNet-152
			& {55.0} & {81.8} & {89.0}
			& {39.4} & {69.2} & {79.1}\\
			NAR~\cite{liu2019neighbor} & ResNet-152
			& {55.1} & {80.3} & {\textbf{89.6}}
			& {39.4} & {68.8} & {79.9}\\
			VSE++~\cite{faghri2017vse} & ResNet-152
			& {52.9} & {80.5} & {87.2}
			& {39.6} & {70.1} & {79.5}\\
			SCO~\cite{huang2018learning} & ResNet-152
			& {55.5} & {\textbf{82.0}} & {89.3}
			& {41.1} & {70.5} & {80.1}\\
			GXN~\cite{gu2018look} & ResNet-152
			& {\textbf{56.8}} & {-} & {\textbf{89.6}}
			& {41.5} & {-} & {80.1}\\
			\midrule
			\textbf{\methiccv} & ResNet-152 & 53.1 & 78.8 & 87.6 & \textbf{42.6} & \textbf{71.6} & \textbf{81.9} \\
			\bottomrule
		\end{tabular}
		\label{tab:flickr1}
		\vspace{-3mm}
	\end{table*}
\end{document}